\documentclass{article}
\usepackage[preprint, nonatbib]{neurips_2023}
\usepackage[utf8]{inputenc} 
\usepackage[T1]{fontenc}    
\usepackage{hyperref}       
\usepackage{url}            
\usepackage{booktabs}       
\usepackage{amsfonts}       
\usepackage{nicefrac}       
\usepackage{microtype}      
\usepackage{xcolor}         
\usepackage{graphicx}
\usepackage{comment}
\usepackage{multicol}
\usepackage{color}
\usepackage{tabularray}
\usepackage{amsmath}
\usepackage{float}
\usepackage{adjustbox}
\usepackage{multirow}

\title{A Benchmark Dataset for Harmful Object Detection}

\author{Eungyeom Ha$^{*1}$ \; Heemook Kim$^{*2}$ \; Sung Chul Hong$^{3}$ \; \stepcounter{footnote}Dongbin Na$^{4}$\thanks{Correspondence to dongbinna@postech.ac.kr} \\ $^1$Yonsei University \; $^2$Inha University \; $^3$Seoul University \; $^4$POSTECH}

\begin{document}

\def\thefootnote{*}\footnotetext{These authors contributed equally to this work.}

\maketitle

\begin{abstract}

Recent multi-media data such as images and videos have been rapidly spread out on various online services such as social network services (SNS).
With the explosive growth of online media services, the number of image content that may harm users is also growing exponentially.
Thus, most recent online platforms such as Facebook and Instagram have adopted content filtering systems to prevent the prevalence of harmful content and reduce the possible risk of adverse effects on users.
Unfortunately, computer vision research on detecting \textit{harmful} content has not yet attracted attention enough.
Users of each platform still manually click the report button to recognize patterns of harmful content they dislike when exposed to harmful content.
However, the problem with manual reporting is that users are already exposed to harmful content.
To address these issues, our research goal in this work is to develop automatic harmful object detection systems for online services.
We present a new benchmark dataset for harmful object detection.
Unlike most related studies focusing on a small subset of object categories, our dataset addresses various categories.
Specifically, our proposed dataset contains more than 10,000 images across 6 categories that might be harmful, consisting of not only normal cases but also \textit{hard cases} that are difficult to detect.
Moreover, we have conducted extensive experiments to evaluate the effectiveness of our proposed dataset.
We have utilized the recently proposed state-of-the-art (SOTA) object detection architectures and demonstrated our proposed dataset can be greatly useful for the real-time harmful object detection task.
The whole source codes and datasets are publicly accessible at \textcolor{blue}{\url{https://github.com/poori-nuna/HOD-Benchmark-Dataset}}.
\end{abstract}

\section{Introduction}

Recently, video and image content has been used in various online services.
However, most online platforms and social media services are still monitoring the uploaded content in a post-processing manner.
The problem with this post-processing approach is that users are exposed to harmful elements, which requires additional costs for organizations.
Thus, recent websites have unprecedentedly needed detection systems for monitoring and regularizing harmful content.
Intelligent cities monitor possibly dangerous objects such as knives and guns in real time~\cite{fernandez2019gun, gonzalez2020real, bhatti2021weapon}.
However, most of the previously presented harmful object detection datasets have limitations in that they address a small subset of harmful object categories or provide only normal cases~\cite{everingham2010pascal, lin2014microsoft, kibria2017analysis}.
In this work, we present a new benchmark dataset for harmful object detection to overcome the limitations of previous studies.
Our dataset contains more than 10,000 images over 6 categories.
The dataset contains guns, knives, and diverse elements like alcohol, insulting gestures, blood, and cigarettes.
The dataset includes normal cases and \textit{hard cases} to detect.
We further explore the effectiveness of our presented dataset and train the state-of-the-art architectures on our introduced dataset.
We demonstrate the trained models achieve modest object detection performance.
Our dataset provides individual \textit{hard case} images, which is greatly useful for evaluating the robustness of harmful object detection algorithms.
For the research purpose, we deploy all the presented datasets, source codes, and trained models.
Our work provides the following main contributions.

\begin{itemize}
  \item We present a novel harmful object detection dataset over 6 categories. To the best of our knowledge, our dataset covers the most various categories compared to the previous studies.
  \item Our dataset includes diverse \textit{hard cases} that are hard to recognize and sometimes induce unexpected detection results, which are useful for evaluating the robustness of detection models.
  \item We publicly provide all the datasets, source codes, and even the trained models for various online media services to utilize our models as off-the-shelf methods easily.
\end{itemize}

\section{Background and Related Work}

\subsection{Object Detection Using Deep Learning}

Deep learning applications in computer vision have garnered significant attention due to their remarkable success across various industry domains~\cite{fernandez2019gun, alzubaidi2021review}.
For image recognition tasks, the recently proposed deep-learning models based on convolutional neural networks (CNN) have shown improved classification performance, surpassing even humans~\cite{wang2020deep, resnet, efficientnet, duan2019centernet, yu2021lite}.
These recently proposed deep CNN architectures suitable for extracting high-level semantic features can be used for various computer vision tasks such as semantic segmentation and object detection~\cite{semanticsegmentation, YOLO, tan2020efficientdet, long2015fully}.
Object detection methods based on R-CNN architectures have been used as baseline object detection models~\cite{girshick2014rich, girshick2015fast, ren2015faster}.
Faster R-CNN has relatively complex architectures. 
However, the detection performance is still competitive compared to the recently proposed methods~\cite{zhu2020deformable, ren2015faster}.
Faster R-CNN generally has been known to show competitive detection performance in that Faster R-CNN results in relatively lower false negatives (FN), although Faster R-CNN is relatively slower~\cite{redmon2018yolov3}.
Therefore, a previous work uses Faster R-CNN as a baseline in FN-critical tasks such as gun and knife detection research~\cite{fernandez2019gun}.
YOLO-based methods have gained popularity for their simplicity and efficacy in real-time processing~\cite{redmon2018yolov3}.
YOLO's streamlined architectures show lower false positives (FP) during real-time detection, suitable for tasks demanding instant feedback~\cite{wang2022fighting}.

\subsection{Harmful Object Detection Dataset}

In the general image object detection research fields, previous studies have presented various image datasets~\cite{deng2009imagenet, everingham2010pascal, lin2014microsoft}.
These datasets provide many image samples with bounding box annotations for various daily objects such as trucks, cars, etc.
Despite its significance, the domain of harmful object detection datasets remains under-explored, specifically given the pressing demand in social media and online live-streaming services.
In particular, online live-streaming service needs to reject harmful object that belongs to harmful categories, such as knives, blood, etc.
Some previous studies have presented harmful object detection datasets~\cite{olmos2018automatic}.
However, most existing studies cover only a subset of the harmful categories or include only easy tasks.
Moreover, the previous studies focus on other types of data, such as chemical signals or sensors, rather than images like ours~\cite{james2014alcohol, al2018alcohol, bhuta2015alcohol, tiwari2018detection, barni2007forensic}.
The details of the studies using other datasets are described in the appendix.
Therefore, we propose a new harmful object detection dataset covering 6 representative harmful categories and \textit{hard cases} with extensive annotation effort of labeler participants.

\section{Proposed Dataset}

\subsection{Category Selection and Annotation Criteria}

Popular online platforms like Instagram, Twitter, and YouTube currently have strict content standards.
However, their standards have tended to be somewhat focused on sexuality.
Some previous studies argue that frequent exposure to violent elements can lead to aggression and desensitization to violence~\cite{rajan2019youth, huesmann2003longitudinal, smith1998harmful}.
Therefore, as preventatives, blocking violent and potentially harmful elements that can lead to unexpected outcomes can be useful for various online platforms.
In addition to violent objects, some studies have observed that visual elements that might be detrimental to users can potentially lead to negative consequences such as addiction and trauma~\cite{grant1997age, cantor1998children, world2008report, bushman2006short, coyne2011luv, anderson2003influence}.
Thus, we have decided to select \textit{alcohol}, \textit{insulting gesture}, \textit{blood}, \textit{cigarette}, \textit{gun}, and \textit{knife} based on a synthesis of prior research, social concerns, and the potential risks associated with exposure to these elements.
A total of 5 labeler participants have collected a dataset of more than 10,000 images using search keywords based on the 6 categories: alcohol, blood, cigarette, gun, insulting gesture, and knife.
A team of three main labelers has gathered over 1,500 images per category.
As we have progressed through the experiments, another 2 labelers have collected additional images of underperforming categories.
We note that each image can have two or more categories in a multi-label classification manner.
The detailed labeling guide for each category is described in the appendix.

\subsection{Data Distribution}

\begin{figure*}[ht]
  \centering
  \centerline{\includegraphics[width=0.9\linewidth]{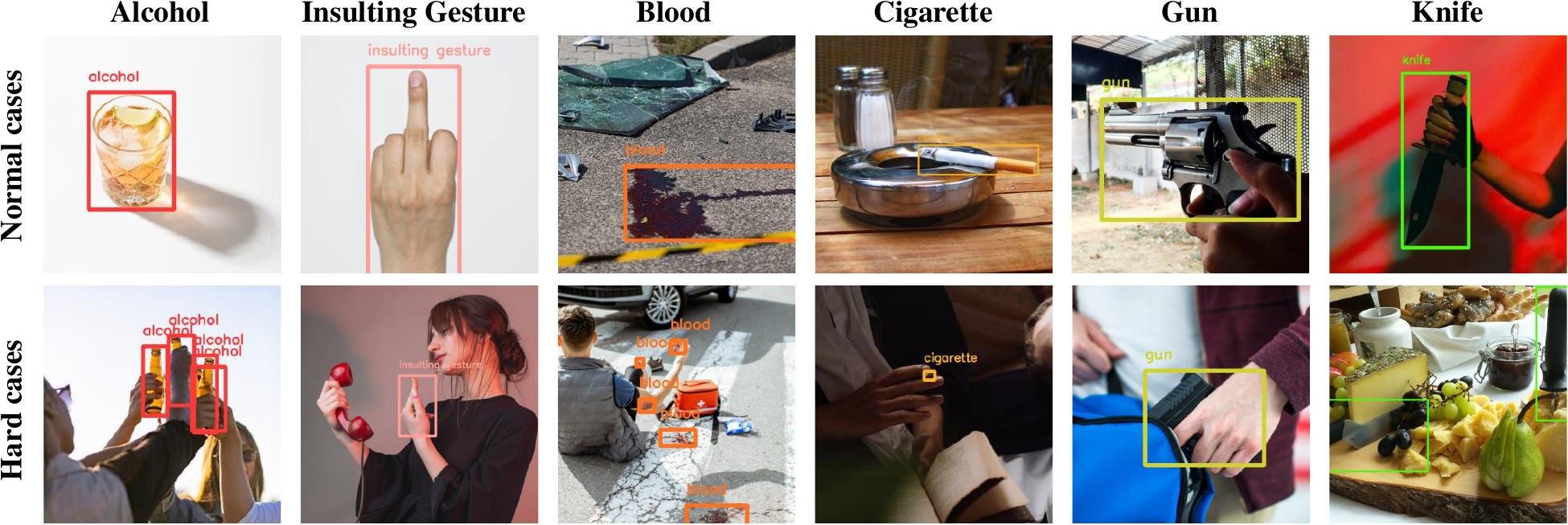}}
  \caption{The example images are randomly sampled from our proposed datasets. 
  The first row shows the normal case images, and the second row shows the \textit{hard case} images.
  The categories denote alcohol, insulting gesture, blood, cigarette, gun, and knife, respectively, in each column.}
  \label{fig: examples for each case}
\end{figure*}

We have divided the dataset into two distinct groups based on the difficulty of detection, the normal cases and the \textit{hard cases}.
The normal cases indicate easily identifiable images.
These images are similar to datasets commonly utilized in existing research.
However, our additional \textit{hard cases} encompass images that are challenging to detect, which is a distinctive contribution from previous studies.
Our \textit{hard case} dataset mainly contains images of harmful objects that are small, or the objects' category-discriminative features are covered.
\textit{Hard cases} also include images that the objects' colors are similar to the background.
Therefore, to infer the label of hard cases, we need other information, such as elements around the object or the context of the image.
More details of the criteria for \textit{hard cases} are demonstrated in the appendix.
After splitting the entire dataset into the normal cases and \textit{hard cases}, we split each dataset again into the training, validation, and test in a ratio of 8:1.5:0.5.
Our extensive efforts ensure the absence of overlapping images between training, validation, and testing datasets with rigorous manual review processes.
The number of data and examples per category for each case is described in Table \ref{tab: Data distribution} and Figure \ref{fig: examples for each case}.

\begin{table}[ht]
\renewcommand\arraystretch{1.0}
\caption{The number of images and instances per category. We note that multiple objects with different categories can belong to an image. 
In those cases, we have counted images that contain multiple categories just once because we have collected the images using search keywords.}
\label{tab: Data distribution}
\begin{adjustbox}{width=14.0cm,center}
\begin{tabular}{c|c|cccccccccccc|cc} \toprule
\multirow{3}{*}{} & \multirow{3}{*}{\textbf{Datasets}} & \multicolumn{14}{c}{\textbf{Categories}} \\ \cline{3-16} \rule{0pt}{11pt}
 &  & \multicolumn{2}{c}{\textbf{Alcohol}} & \multicolumn{2}{c}{\textbf{Insulting Gesture}} & \multicolumn{2}{c}{\textbf{Blood}} & \multicolumn{2}{c}{\textbf{Cigarette}} & \multicolumn{2}{c}{\textbf{Gun}} & \multicolumn{2}{c|}{\textbf{Knife}} & \multicolumn{2}{c}{\textbf{All}} \\
 &  & Images & Instances & Images & Instances & Images & Instances & Images & Instances & Images & Instances & Images & Instances & Images & Instances \\ \midrule
\multirow{4}{*}{\textbf{Normal Cases}} 
 & Train    & 453 & 453 & 396 & 396 & 470 & 470 & 467 & 467 & 849 & 849 & 2011 & 2011 & 4646 & 4646 \\
 & Valid    &  54 &  54 &  47 &  47 &  57 &  57 &  56 &  56 & 101 & 101 &  237 &  237 &  552 &  552 \\
 & Test     &  26 &  26 &  23 &  23 &  27 &  27 &  27 &  27 &  49 &  49 &  118 &  118 &  270 &  270 \\ \cline{2-16} \rule{0pt}{11pt}
 & Subtotal & 533 & 533 & 466 & 466 & 554 & 554 & 550 & 550 & 999 & 999 & 2366 & 2366 & 5468 & 5468 \\ \midrule
\multirow{4}{*}{\textbf{Hard Cases}} 
 & Train    & 831 & 3013 & 226 & 404 & 844 & 2525 & 1307 & 3942 & 481 & 744 & 697 & 1242 & 4386 & 11870 \\
 & Valid    &  99 &  367 &  28 &  50 & 101 &  317 &  155 &  512 &  57 &  81 &  82 &  151 &  522 &  1478 \\
 & Test     &  48 &  234 &  13 &  28 &  49 &  137 &   76 &  300 &  28 &  48 &  41 &   77 &  255 &   824 \\ \cline{2-16} \rule{0pt}{11pt}
 & Subtotal & 978 & 3614 & 267 & 482 & 994 & 2979 & 1538 & 4754 & 566 & 873 & 820 & 1470 & 5163 & 14172 \\ \midrule
 & \textbf{Total} & \textbf{1511} & \textbf{4147} & \textbf{733} & \textbf{948} & \textbf{1548} & \textbf{3533} & \textbf{2088} & \textbf{5304} & \textbf{1552} & \textbf{1872} & \textbf{3186} & \textbf{3836} & \textbf{10631} & \textbf{19640} \\ 
 \bottomrule
\end{tabular}
\end{adjustbox}
\end{table}

\section{Experiments}

We have utilized two baseline object detection architectures, YOLOv5~\cite{YOLO} and Faster R-CNN~\cite{ren2015faster}, which are representative one-stage and two-stage object detection methods, respectively.
When reporting the main experimental results, we have adopted consistent hyperparameter settings for object detection models to obtain reliable and reproducible results.

\subsection{Hyperparameter Tuning and Model Optimization}

We have extensively experimented with various hyperparameters, such as batch size and weight initialization methods. For training YOLOv5 models, we use an image size of 416 and a batch size of 32. We have observed that the detection performance of the YOLOv5 models has converged after 200 epochs. For training Faster R-CNN models, we have adopted the MMDetection framework~\cite{chen2019mmdetection}, which has been notably known to provide baseline benchmarks in object detection research fields. We have experimented with various hyperparameter settings to fully leverage the detection framework. We have trained the Faster R-CNN models for 150 epochs with a learning rate of 0.0025, which demonstrates competitive detection performance in our harmful object detection tasks.

\subsection{Model Training and Evaluation}

We have trained object detection models and evaluated their effectiveness in various scenarios.
Specifically, we have trained YOLOv5 and Faster R-CNN models using two distinct dataset configurations. 
The first setting contains only normal cases in the training dataset, while the second setting consists of normal and \textit{hard cases} in the training dataset.
Formally, in the first setting, we train the object detection models on only the normal case training dataset $\mathcal{D}_{normal}^{train}$ and evaluate their detection performance on the normal case test dataset $\mathcal{D}_{normal}^{test}$ and the hard case test dataset $\mathcal{D}_{hard}^{test}$ individually. In the second setting, we train the object detection models on the joint training data distribution ${\mathcal{D}_{normal}^{train} \cup \mathcal{D}_{hard}^{train}}$ that consists of normal and \textit{hard cases} during the training time.
We expect that the second setting shows achieve improved generalization performance, which is desirable in that the object detection model trained on ${\mathcal{D}_{normal}^{train} \cup \mathcal{D}_{hard}^{train}}$ can capture more abundant feature representations of various difficult objects compared to models trained on the only normal case training dataset.

\subsection{Performance Metrics}

We have evaluated our object detection models using mAP (Mean Average Precision). This mAP metric measures the comprehensive detection performance of an object detection model by calculating precision scores at different recall levels. Specifically, we have adopted 2 representative variations of the mAP: mAP@50-95 and mAP@95. 
We have comprehensively considered the various IoU thresholds and reported the overall detection results of trained models across 6 categories.
We note that the mAP scores can be individually calculated at each different confidence threshold during evaluations. 
Therefore, we have thoughtfully selected the confidence thresholds that produce the highest mAP scores for each model utilizing the validation datasets.
For YOLOv5 models, the optimal confidence score is 0.3, resulting in the highest mAP score for the test dataset.
Meanwhile, for the Faster R-CNN models, the best confidence score is calculated as 0.1. 
We have meticulously assessed and compared the performance of our models~\cite{lin2014microsoft, huang2017speed} by reviewing the confidence thresholds and mAP scores.
We hope that our experimental results provide valuable insights into the object detection research fields, leading to especially contribute to the advancement of harmful object detection research.

\begin{table}[ht]
\renewcommand\arraystretch{1.0}
\caption{Detection performance of trained models. The table shows mAP scores per category.}
\label{tab: Detection performance}
\begin{adjustbox}{width=14.0cm,center}
\begin{tabular}{c|c|c|c|cccccc|c} \toprule
\multirow{2}{*}{\textbf{Train Dataset $\mathcal{D}^{train}$}}& \multirow{2}{*}{\textbf{Test Datset $\mathcal{D}^{test}$}} & \multirow{2}{*}{\textbf{Models}} & \multirow{2}{*}{\textbf{Performance Measures}}& \multicolumn{7}{c}{\textbf{Categories}}\\ \cline{5-11} \rule{0pt}{11pt}
 & & & & {\textbf{Alcohol}} & {\textbf{Insulting Gesture}} & {\textbf{Blood}} & {\textbf{Cigarette}} & {\textbf{Gun}} & {\textbf{Knife}} & {\textbf{All}} \\ \midrule 
\multirow{8}{*}{\scalebox{1.5}{\textbf{$\mathcal{D}^{train}_{normal}$}}}
 & \multirow{4}{*}{\scalebox{1.5}{\textbf{$\mathcal{D}^{test}_{normal}$}}} & \multirow{2}{*}{\textbf{YOLOv5}} & mAP@50    & 97.4 & 97.8 & 69.8 & 89.2 & 91.6 & 95.0 & 90.1 \\
 & &                                                                                                          & mAP@50-95 & 89.7 & 85.1 & 48.6 & 79.9 & 75.1 & 80.5 & 76.5 \\ \cline{3-11} \rule{0pt}{11pt}
 & & \multirow{2}{*}{\textbf{Faster R-CNN}}                                                                   & mAP@50    & 89.3 & 99.3 & 73.6 & 83.5 & 90.0 & 87.7 & 87.2 \\
 & &                                                                                                          & mAP@50-95 & 72.3 & 76.9 & 39.9 & 60.5 & 62.5 & 62.2 & 62.4 \\ \cline{2-11} \rule{0pt}{11pt}
& \multirow{4}{*}{\scalebox{1.5}{\textbf{$\mathcal{D}^{test}_{hard}$}}} & \multirow{2}{*}{\textbf{YOLOv5}}    & mAP@50    & 55.2 & 66.7 & 44.7 & 41.2 & 61.5 & 49.7 & 53.2 \\
 & &                                                                                                          & mAP@50-95 & 40.2 & 47.9 & 27.7 & 26.2 & 43.3 & 39.4 & 37.4 \\ \cline{3-11} \rule{0pt}{11pt}
 & & \multirow{2}{*}{\textbf{Faster R-CNN}}                                                                   & mAP@50    & 39.0 & 55.2 & 20.1 & 10.7 & 33.9 & 38.0 & 32.8 \\
 & &                                                                                                          & mAP@50-95 & 24.5 & 38.5 &  9.4 &  4.8 & 22.0 & 19.3 & 19.8 \\ \midrule
\multirow{8}{*}{\scalebox{1.5}{\textbf{${\mathcal{D}_{normal}^{train} \cup \mathcal{D}_{hard}^{train}}$}}}
 & \multirow{4}{*}{\scalebox{1.5}{\textbf{$\mathcal{D}^{test}_{normal}$}}} & \multirow{2}{*}{\textbf{YOLOv5}} & mAP@50    & 99.2 & 99.5 & 79.1 & 95.5 & 98.4 & 95.1 & 94.5 \\
 & &                                                                                                          & mAP@50-95 & 92.8 & 87.4 & 58.4 & 80.2 & 86.9 & 83.2 & 81.5 \\ \cline{3-11} \rule{0pt}{11pt}
 & & \multirow{2}{*}{\textbf{Faster R-CNN}}                                                                   & mAP@50    & 96.1 & 100.0& 66.0 & 85.0 & 94.3 & 89.7 & 88.5    \\
 & &                                                                                                          & mAP@50-95 & 79.5 & 74.9 & 39.1 & 62.2 & 64.5 & 61.7 & 63.6 \\ \cline{2-11} \rule{0pt}{11pt}
& \multirow{4}{*}{\scalebox{1.5}{\textbf{$\mathcal{D}^{test}_{hard}$}}} & \multirow{2}{*}{\textbf{YOLOv5}}    & mAP@50    & 91.9 & 75.5 & 70.2 & 88.2 & 76.2 & 74.9 & 79.5\\
 & &                                                                                                          & mAP50-95  & 75.7 & 57.3 & 46.8 & 63.1 & 59.5 & 55.4 & 59.6 \\ \cline{3-11} \rule{0pt}{11pt}
 & & \multirow{2}{*}{\textbf{Faster R-CNN}}                                                                   & mAP@50    & 83.1 & 64.7 & 57.4 & 78.2 & 64.9 & 52.8 & 66.9  \\
 & &                                                                                                          & mAP@50-95 & 57.9 & 41.4 & 28.7 & 45.8 & 36.6 & 31.9 & 40.4 \\ \bottomrule
\end{tabular}
\end{adjustbox}
\end{table}

\subsection{Analysis of Experimental Results}

We note that the following four datasets do not overlap with each other.
For training the models, we provide (1) normal case training dataset $\mathcal{D}_{normal}^{train}$ and (2) \textit{hard case} training dataset $\mathcal{D}_{hard}^{train}$.
For testing the models, we provide (3) normal case test dataset $\mathcal{D}_{normal}^{test}$, (4) \textit{hard case} test dataset $\mathcal{D}_{hard}^{test}$.

\begin{figure}[ht]
  \centering
  \centerline{\includegraphics[width=0.8\linewidth]{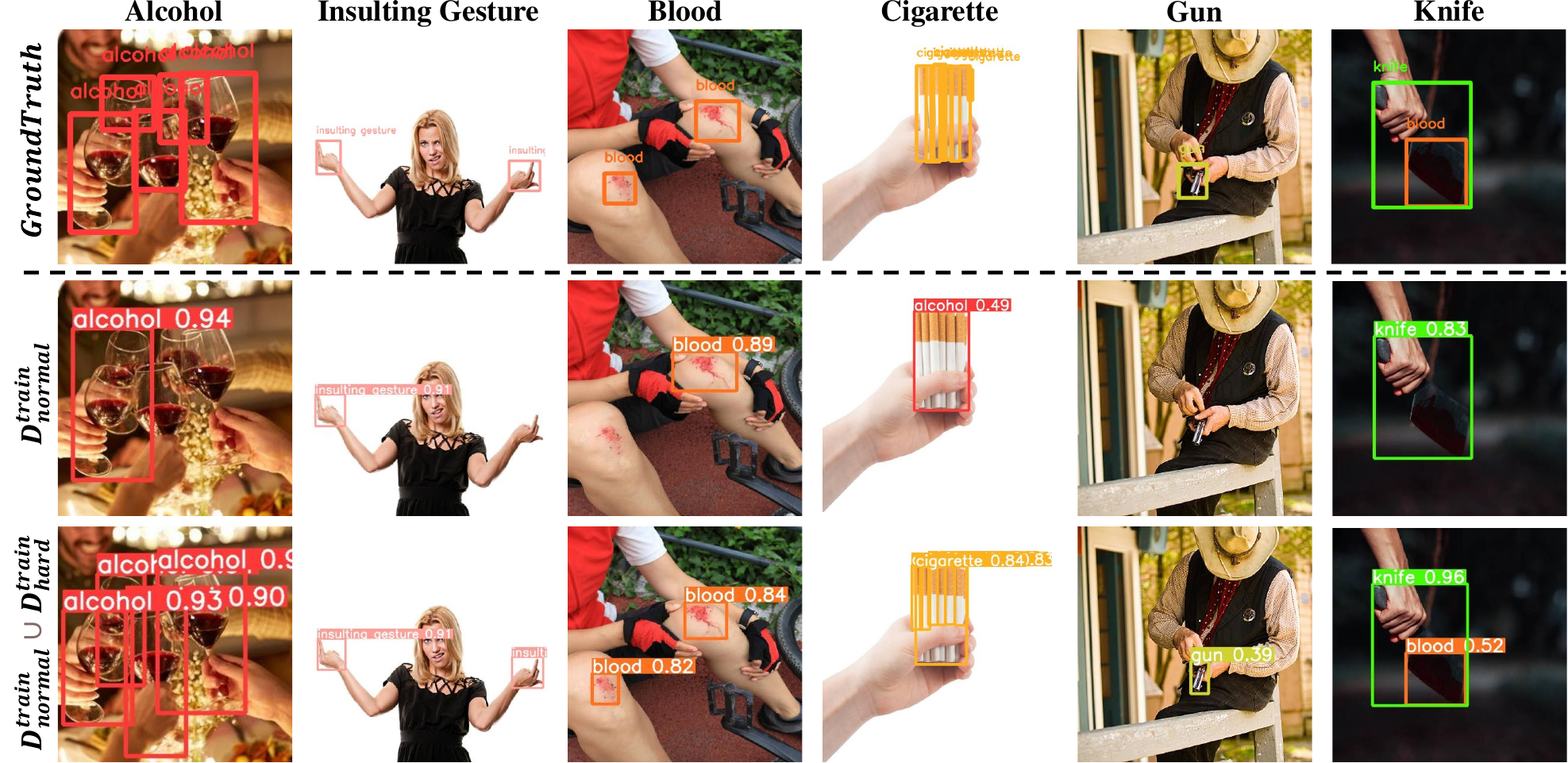}}
  \caption{
  The example images from our \textit{hard case} test dataset and the corresponding inference results. The first row represents \textit{hard case} ground-truth samples from the dataset $\mathcal{D}_{hard}^{test}$. The second row shows the detection results using YOLOv5 trained on only $\mathcal{D}_{normal}^{train}$. The third row represents the detection results using YOLOv5 trained on the joint dataset ${\mathcal{D}_{normal}^{train} \cup \mathcal{D}_{hard}^{train}}$. We have found \textit{hard case} training images can be greatly useful to improve the robustness of the detection models.}
  \label{fig: hardcases results YOLO}
\end{figure}

The main results of the experiments are presented in Table \ref{tab: Detection performance}. We note that training the detection models on the joint dataset $\mathcal{D}_{normal}^{train} \cup \mathcal{D}_{hard}^{train}$ that contains normal and \textit{hard cases} improves the overall detection performance across whole categories. The YOLOv5 models have improved the average mAP by 5, from 76.5 to 81.5, on the normal case test dataset $\mathcal{D}_{normal}^{test}$. The category with the largest performance improvement is \textit{gun}. The detection performance has been improved by 11.8, from 75.1 to 86.9. We note that the YOLOv5 model also shows significantly improved detection performance across whole categories by 22.2, from 37.4 to 59.6, on the \textit{hard case} test dataset $\mathcal{D}_{hard}^{test}$. The \textit{cigarette} category shows the largest performance improvement. We have found that the detection performance has been improved by 36.9, from 26.2 to 63.1.
We have observed that the Faster R-CNN model also achieves an improvement of the detection overall performance by 20.6, from 19.8 to 40.4, on the \textit{hard case} test dataset $\mathcal{D}_{hard}^{test}$ when using $\mathcal{D}_{normal}^{train} \cup \mathcal{D}_{hard}^{train}$. The largest performance gain occurs in the \textit{cigarette} category, improved by 41, from 4.8 to 45.8. The examples of inference results on some \textit{hard case} test samples are illustrated in Figure \ref{fig: hardcases results YOLO}.
Our experiments show \textit{hard case} training dataset is crucial to achieving robust detection performance on various difficult objects.

\section{Conclusion}

In this paper, we present a new benchmark dataset that is useful for the harmful object detection task, helping various users and organizations automatically address potentially harmful elements of visual content. We provide datasets that cover 6 categories: alcohol, blood, cigarette, gun, insulting gesture, and knife. For constructing the harmful object detection dataset, we have first chosen normal cases of images similar to the datasets used in previous studies. However, we note that the \textit{hard cases} that are hardly recognizable frequently induce unexpected model outputs. With extensive experiments, we have demonstrated training the detection model on the normal and \textit{hard cases} simultaneously shows improved detection performance over all 6 categories. The experimental results conclude that \textit{hard cases} training samples are greatly useful for recognizing various shapes of harmful objects. We hope our presented datasets, trained models, and source codes can be utilized for various online services and research fields that adopt visual censorship systems.

\bibliographystyle{plain}
\bibliography{neurips_2023}

\newpage

\appendix

\section{Analysis of Existing Harmful Object Detection Datasets}

\begin{itemize}
    \item \textbf{Alcohol}: Most existing alcohol detection work has been concerned with drunk driving~\cite{james2014alcohol, al2018alcohol, bhuta2015alcohol}. These previous studies have not mainly aimed to detect alcohol images but to recognize the human body's chemical signals or physical reactions after drinking alcohol. In the real world, alcohol can lead to unexpected accidents due to drunk driving. Therefore, we adopt the alcohol object category for our dataset. Likewise, avoiding frequent exposure to alcohol on the Internet can be considered important as a precaution.
    \item \textbf{Blood}: Existing studies on blood detection are mainly based on medical or forensic perspectives~\cite{tiwari2018detection, barni2007forensic}. The previous datasets are frequently used to detect diseases through blood tests or to detect bloodstains on crime scenes and evidence using the luminol chemical reaction. Therefore, the existing work either has dealt with blood images on a cellular level or uses data from chemical sensors. The bloodstain images themselves have not been frequently treated as datasets.
    \item \textbf{Cigarette}: The purpose of traditional cigarette detection studies is to prevent risky incidents caused by smoking behavior. Their data samples generally include fire signals caused by smoking at gas stations and accidents caused by smoking while driving~\cite{9816219, kharade2021image, chang2019dangerous}. Therefore, many studies have focused on detecting cigarette smoke. Different from previous studies, we aim to detect the cigarette object itself.
    \item \textbf{Sign language}: The existing finger pose research addresses sign language detection for hearing-impaired people~\cite{moryossef2020real, shukor2015new}. Their work has focused to detect and interpret what the sign language represents based on estimating the pose of a person's fingers. They generally utilize tilt and accelerometer sensors on the fingers through data gloves. However, most of the existing studies do not deal with images of insulting hand gestures used by hearing-abled people. To the best of our knowledge, we are the first to address and provide a dataset of insulting hand gestures in images.
    \item \textbf{Weapon}: Existing studies on harmful elements mainly focus on weapons such as guns and knives~\cite{narejo2021weapon, ingle2022real}. Unlike our study, previous studies generally have intended to prevent real-world violence and terrorism, not to detect harmful elements in internet content. Gun and knife detection is necessary to prevent terrorism in the real world. Therefore, we also adopt these categories, yet, focus on the Internet media content to collect images. We note that many studies claim that frequent exposure to weapons such as guns and knives leads to familiarity with them~\cite{bushman2006short, anderson2003influence}. Thus, we also consider these categories as harmful categories on the Internet as a precaution.
\end{itemize}

\section{Rationale for Category Selection}

\begin{itemize}
  \item \textbf{Alcohol}: Numerous studies have highlighted the risks of early alcohol exposure, suggesting that exposure to alcohol objects can pave the way for substance misuse disorders in later life~\cite{grant1997age}. By identifying and moderating such content, we aim to mitigate the normalization of underage excessive alcohol consumption.
  \item \textbf{Blood}: Graphic visuals, particularly those displaying blood or gore, can sometimes induce fear and trauma in younger audiences~\cite{cantor1998children}. Restricting access to such visuals may aid in fostering a safer media environment for children and adolescents.
  \item \textbf{Cigarette}: The World Health Organization (WHO) has persistently warned about the dangers of youth tobacco consumption, indicating that early exposure can lead to lifelong addiction~\cite{world2008report}. By detecting and blurring such content, the allure and curiosity surrounding smoking might be reduced.
  \item \textbf{Gun}: Numerous studies have pointed out that exposure to firearms in media can influence aggressive behaviors and desensitize youth to real-life violence~\cite{bushman2006short}. Therefore, by moderating this content, the objective diminishes the potential for gun-related curiosities and imitative behaviors.
  \item \textbf{Insulting Gesture}: Exposure to inappropriate or obscene gestures can mold the negative social behaviors of youngsters, often leading to the replication of such gestures in inappropriate situations~\cite{coyne2011luv}. The identification of these gestures aims to cultivate better behavioral norms.
  \item \textbf{Knife}: The representation of weapons, especially sharp ones like knives, has been correlated with an increased propensity for violent behaviors in young individuals~\cite{anderson2003influence}. Preventing young audiences from these visual triggers can potentially curtail the glamorization of violence.
\end{itemize}

\section{Sources of Images and Labeling Tools}

Our team of labelers has crawled images from the following websites in which users can use photos for research purposes.

\begin{itemize}
    \item \url{https://pexels.com}
    \item \url{https://unsplash.com}
    \item \url{https://freeimages.com}
    \item \url{https://freepik.com}
    \item \url{https://pixabay.com}
    \item \url{https://flickr.com}
    \item \url{https://istockphoto.com}
\end{itemize}

\begin{figure}[H]
\centering
\includegraphics[width=0.9\linewidth]{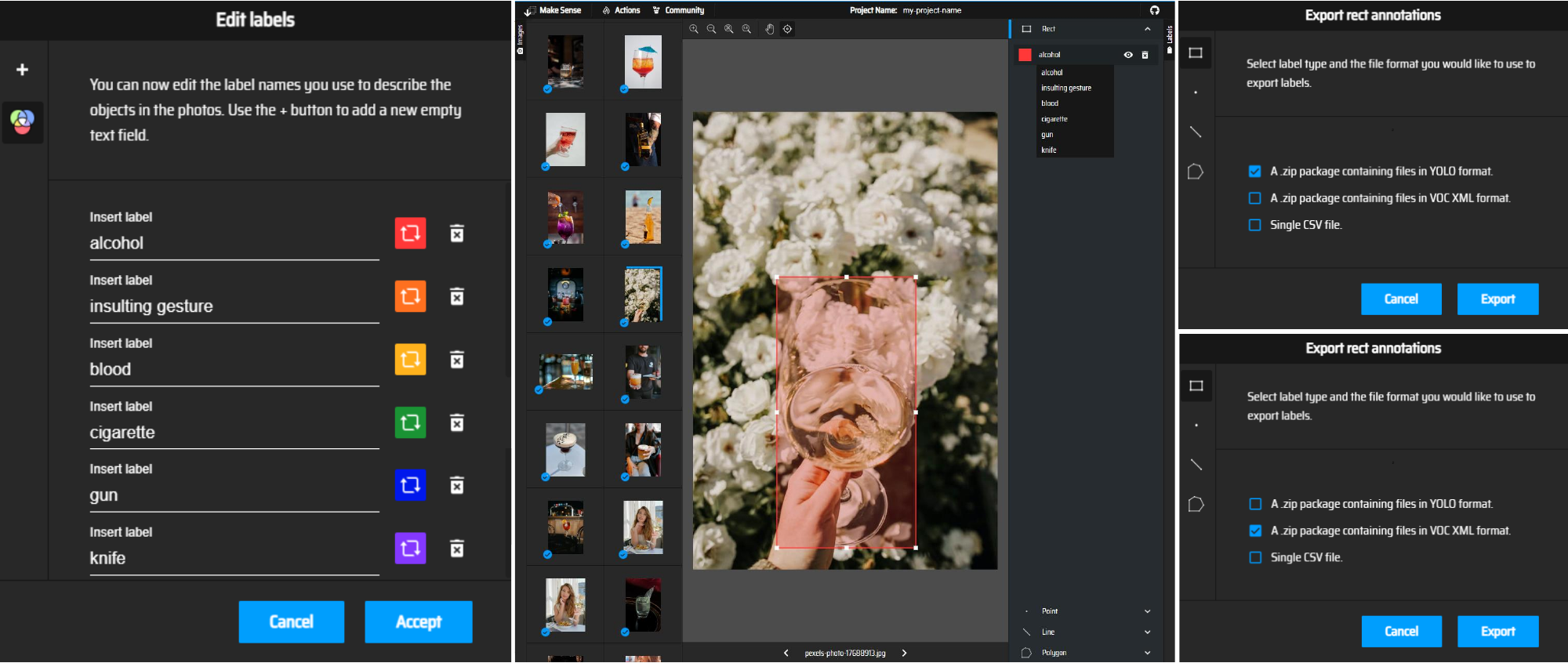}
\caption{The illustration of the labeling procedure. The captured screenshots represent the process of annotating and labeling with the \textit{Makes Sense} annotation tool.}
\label{fig: makesense}
\end{figure}

Labeler participants have utilized \href{https://chrome.google.com/webstore/detail/one-click-image-downloade/djcobamaplcmhmaocomnkfdbcoiggepo}{One Click Image Downloader} for crawling and \href{https://www.makesense.ai}{Make Sense} for labeling. Labelers have set the whole categories first, then proceeded to annotate and label all images. Labelers have utilized rectangular bounding boxes for the annotation. After the labelers finished labeling, they exported annotation files in YOLO and VOC formats. The actual procedure conducted by the labelers is illustrated in Figure \ref{fig: makesense}.

\section{Labeling Guidline}

\begin{itemize}
  \item \textbf{Alcohol}: Any bottle that could be recognized as a bottle of alcohol, no matter its shape, labelers have labeled the object as alcohol. We note that labelers have labeled not only the bottles but also any glasses that can be recognized as alcoholic beverages by the labelers.
  \item \textbf{Blood}: Labelers have labeled the objects as blood if blood is on the objects. Moreover, if blood is widely scattered, the labelers have grouped the blood regions and labeled them as a single blood object.
  \item \textbf{Cigarette}: Labelers have labeled all the individual cigarettes as possible. The labelers also have labeled e-cigarettes as cigarettes. Moreover, labelers have labeled cigarette packs as cigarettes.
  \item \textbf{Gun}: If labelers could recognize the object as a gun, whether the shape is a pistol, rifle, or sniper rifle, they have labeled the objects as guns.
  \item \textbf{Insulting Gesture}: The insulting gestures can appear differently according to the country. Therefore, after extensive discussion, the labelers agreed to define a specific finger shape commonly used worldwide as an insulting gesture. Specifically, the labelers have annotated the hand with only the middle finger extended and the rest of the fingers folded as an insulting gesture.
  \item \textbf{Knife}: Regardless of the type of knife, such as kitchen knives and long swords, labelers have labeled the objects as knives.
\end{itemize}

\begin{figure}[H]
  \centering
  \includegraphics[width=0.5\linewidth]{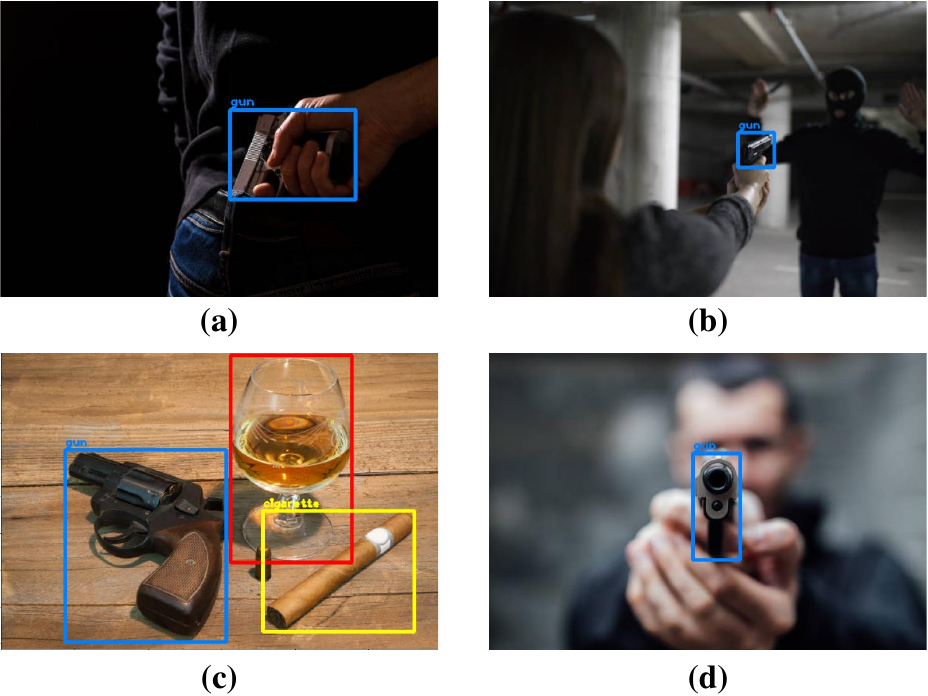}
  \caption{Representative image examples showing how the \textit{hard case} criteria are applied. (a) The image of a gun with a concealed barrel. (b) The image of a gun with both the width and height is less than 0.2 relative to the image size. (c) The image contains alcohol, a cigarette, and a gun simultaneously in a multi-label manner. (d) The image of a gun is viewed from the direction of the front barrel, which is hard to recognize.}
  \label{fig: hardcase_example_gun}
\end{figure}

\section{Hard Case Criteria}

As we have noted, our hard cases contain images that are hard to detect.
The criteria we have set are as follows.
(1) Images that contain various objects with different categories in a multi-label manner.
(2) Images that have small-sized harmful objects.
The exact criteria for size are that both the width and height of the harmful objects relative to the image's size are less than 0.2.
(3) Images that have harmful objects whose category-discriminative features are unrecognizable.
These cases include when the object's color is similar to the background, the object is viewed from an unusual angle, or the category-discriminative features of the object are concealed.
Additional criteria for each category are as follows.

\begin{itemize}
  \item \textbf{Alcohol}: Images that are taken from above or below, rather than from the side, so that the shape of the bottle or glass can not be distinguishable.
  \item \textbf{Blood}: We note that the golden standard criteria of blood itself can be ambiguous. For example, scattered blood is grouped and labeled by the subjectivity of labelers. We have classified the images with multiple blood groups as hard cases.
  \item \textbf{Cigarette}: Images that the object has a relatively different shape, not a single cigarette, such as cigarette cases or electronic cigarettes.
  \item \textbf{Gun}: Images people can recognize as a gun, even though the barrel is heavily concealed. Additionally, we also consider the images of a gun that are viewed from the direction of the front barrel as a \textit{gun}.
  \item \textbf{Insulting gesture}: Images taken from the side of the hand. Therefore, it is hard to tell if it is the middle finger.
  \item \textbf{Knife}: Images humans could recognize as a knife, even though the blades are not visible and only the handle is visible.
\end{itemize}

\begin{figure}[H]
  \centering
  \centerline{\includegraphics[width=0.9\linewidth]{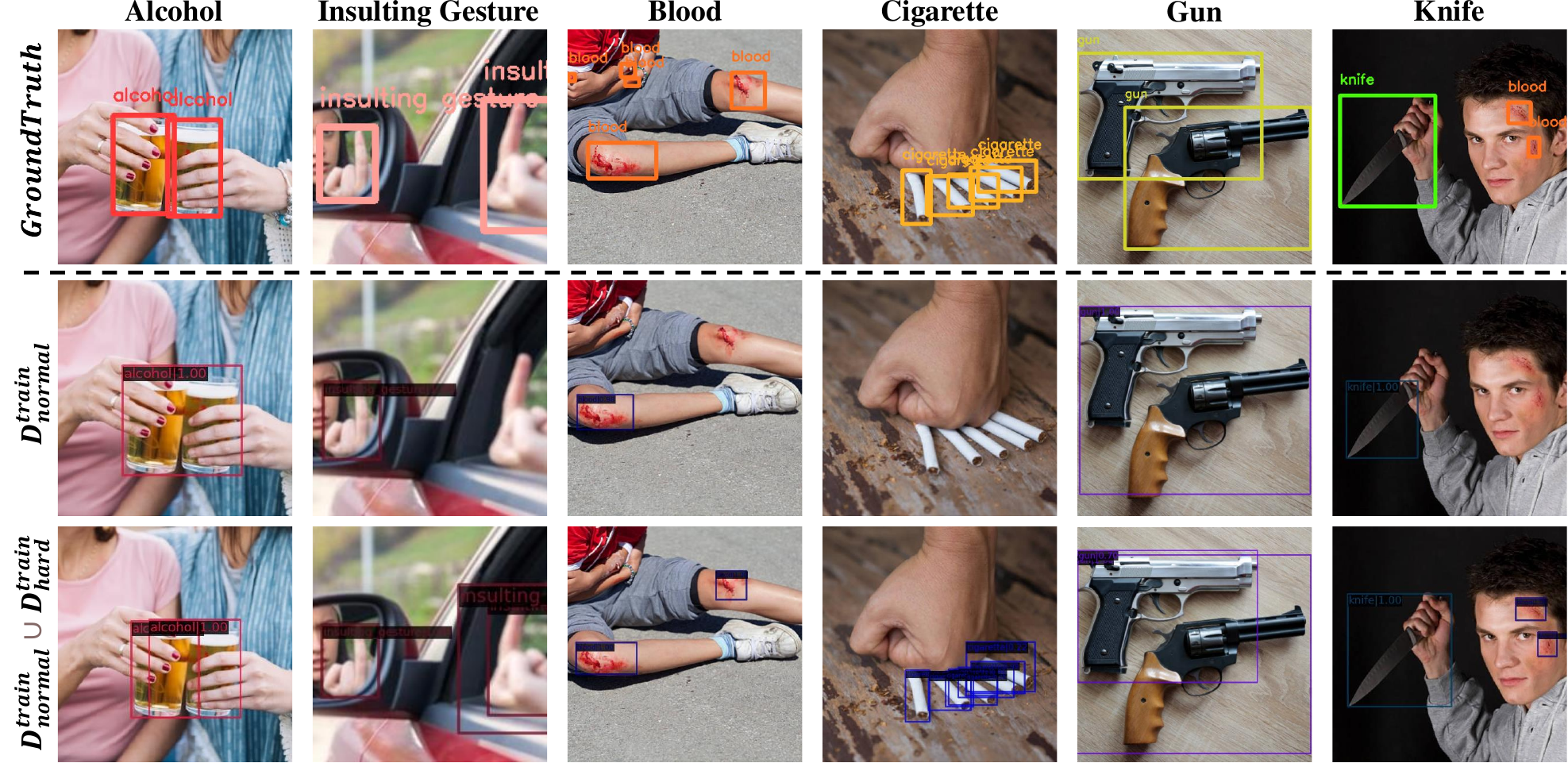}}
  \caption{The example images from our hard case test dataset and the corresponding inference results. The first row represents hard case ground-truth samples from the dataset $\mathcal{D}_{hard}^{test}$. The second row shows the detection results using Faster R-CNN trained on only $\mathcal{D}_{normal}^{train}$. The third row represents the detection results using Faster R-CNN trained on the joint dataset ${\mathcal{D}_{normal}^{train} \cup \mathcal{D}_{hard}^{train}}$. We have found hard case training images can greatly improve the detection models' robustness.}
  \label{fig: rcnn hardcases}
\end{figure}

The representative images of \textit{hard cases} our criteria apply are described in Figure \ref{fig: hardcase_example_gun}. Moreover, we show the ground-truth image samples and the object detection results using the Faster R-CNN models in Figure~\ref{fig: rcnn hardcases}. Similar to the YOLOv5 models, the Faster R-CNN models demonstrate improved object detection performance when using the \textit{hard case} training dataset.

\section{Model Architectures}

We have utilized two baseline object detection architectures, YOLOv5~\cite{YOLO} and Faster R-CNN~\cite{ren2015faster}, which are representative one-stage and two-stage object detection methods, respectively.

\begin{figure}[H]
\centering
\includegraphics[width=0.9\linewidth]{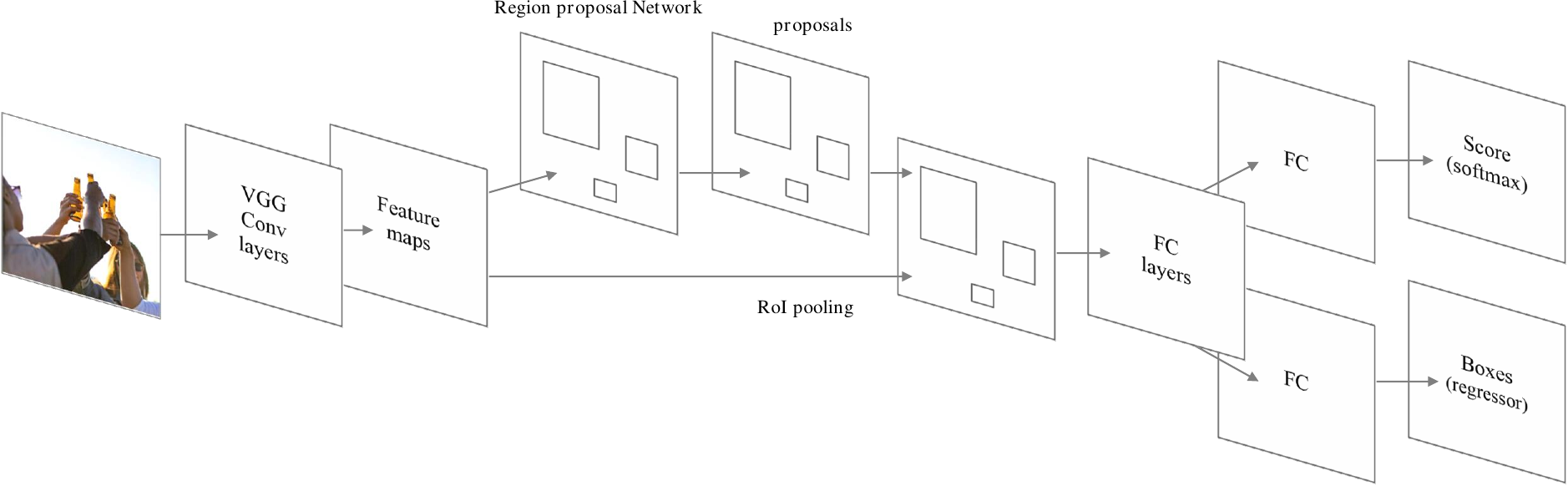}
\caption{The illustration of the Faster R-CNN network architecture.}
\label{fig: faster rcnn architecture}
\end{figure}

\noindent Faster R-CNN, with the Region Proposal Network (RPN), is a highly precise object detection method~\cite{ren2015faster}. 
The approach comprises two stages, with the first stage utilizing a convolutional layer to extract features and generate feature maps. 
The region candidate network then creates candidate boxes, while the region of interest pooling layer collects feature maps and regional candidate frames. 
In the final stage, the classification layer identifies the object category and adjusts the position of the candidate frames. 
The VGG16 architecture can be utilized to detect small targets due to the low-resolution representations that are down-sampled and small pixel sizes on the feature space~\cite{simonyan2014very}.
The structure of Faster R-CNN is described in Figure \ref{fig: faster rcnn architecture}.

\begin{figure}[H]
\centering
\includegraphics[width=0.9\linewidth]{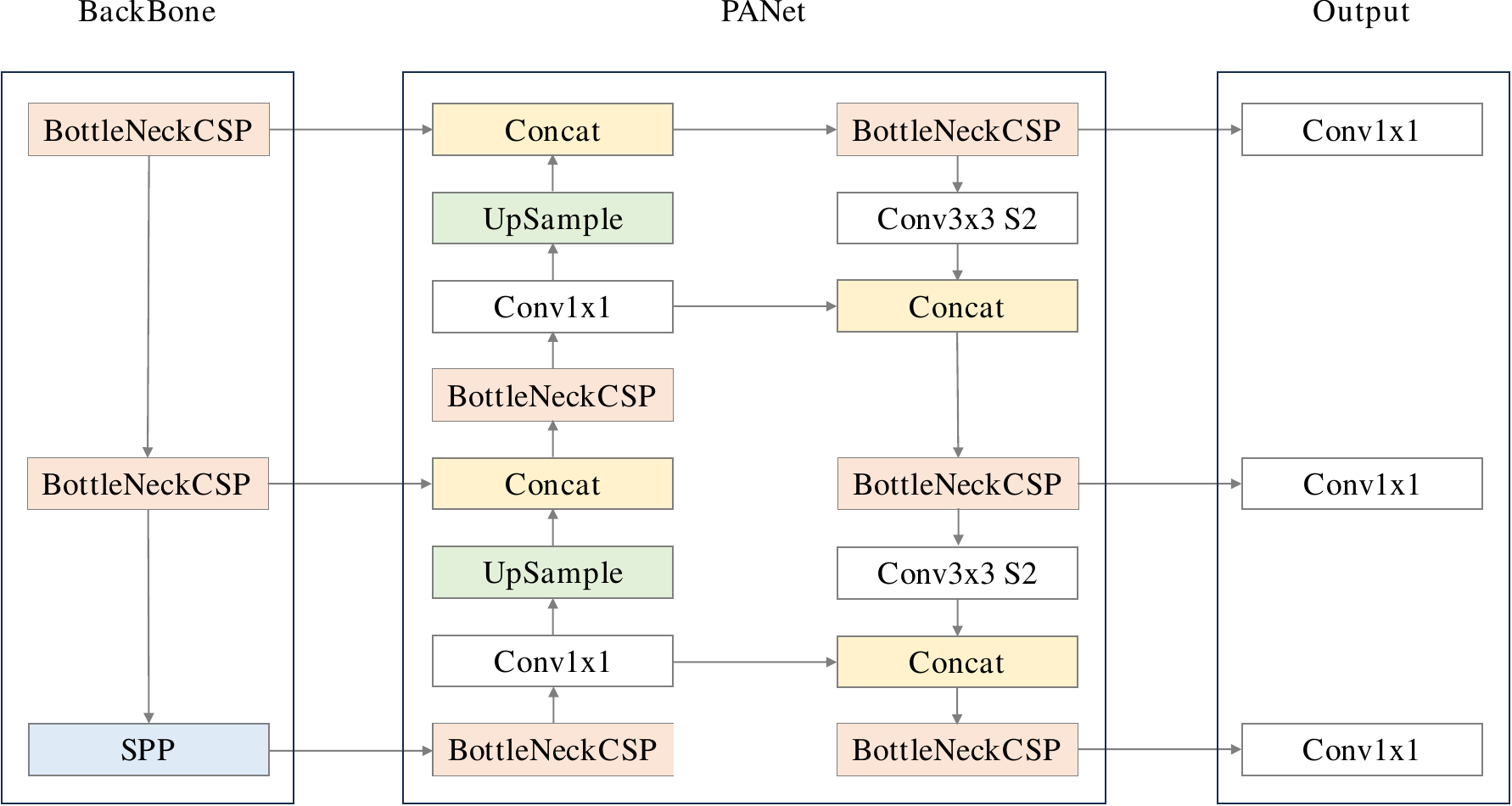}
\caption{The illustration of the YOLOv5 network architecture.}
\label{fig: yolov5 architecture}
\end{figure}

\noindent YOLOv5, which stands for "You Only Look Once," is a one-stage, regression-based method for real-time object detection~\cite{YOLO}. 
It offers end-to-end training, determining the target category and positioning simultaneously. 
The network structure consists of only convolutional layers and the input image layer. The YOLOv5 has been known for its lightweight and quick detection performance, surpassing other methods like Faster R-CNN in speed and precision benchmarks~\cite{bochkovskiy2020yolov4, redmon2018yolov3}. The architecture of YOLOv5 is described in Figure \ref{fig: yolov5 architecture}.

\begin{figure}[ht]
\centering
\includegraphics[width=0.5\linewidth]{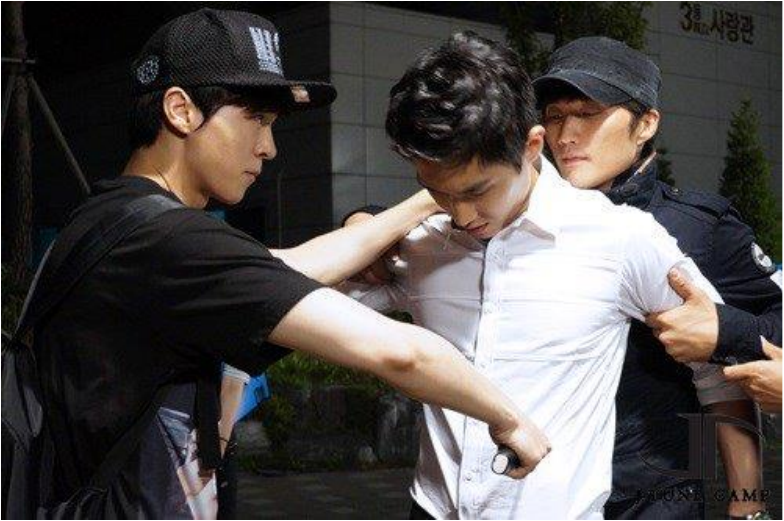}
\caption{An image example of a man stabbing another man with a knife.}
\label{fig: hardcase}
\end{figure}

\section{Discussion}

Our goal is to develop an automated classification system to detect potentially harmful objects and prevent exposure to the harmful objects.
Therefore, the detection model should be able to detect harmful objects in various cases, including normal and \textit{hard cases}.
The final goal is to train many \textit{hard cases} so that models can detect harmful elements like humans, even when the distinguishing features of harmful objects are concealed.
The ideal object detection model can detect harmful objects based on the overall context of the images, even if the most significant features of the harmful elements are masked.
Figure \ref{fig: hardcase} is an example of a knife in which the blade of the knife is hidden.
This example is a representative scene that needs to be detected as a knife. 
However, the \textit{hard case} objects can not be easily detected when training object detection models only on normal case images.
Thus, we will continue collecting, training, and deploying additional hard cases to detect harmful objects, even in hard cases like Figure~\ref{fig: hardcase} for future work.

\end{document}